\documentclass[conference]{IEEEtran}
\IEEEoverridecommandlockouts

\usepackage{cite}
\usepackage{bm}
\usepackage{amsmath,amssymb,amsfonts}
\usepackage{graphicx}
\usepackage{textcomp}
\usepackage{xcolor}
\usepackage{booktabs}
\usepackage{caption}
\usepackage{subfigure}
\usepackage{algorithm}
\usepackage{algpseudocode}
\usepackage{multirow}
\usepackage{longtable}
\usepackage{url}

\def\BibTeX{{\rm B\kern-.05em{\sc i\kern-.025em b}\kern-.08em
    T\kern-.1667em\lower.7ex\hbox{E}\kern-.125emX}}
\begin{document}

\title{Large Language Model Assisted Adversarial Robustness Neural Architecture Search \\}

\author{\IEEEauthorblockN{Rui Zhong$^*$, Yang Cao}
\IEEEauthorblockA{\textit{Graduate School of Information Science and Technology} \\
\textit{Hokkaido University}\\
Sapporo, Japan \\
\{rui.zhong.u5, yang.cao.y4\}@elms.hokudai.ac.jp}
\and 
\IEEEauthorblockN{Jun Yu}
\IEEEauthorblockA{\textit{Institute of Science and Technology} \\
\textit{Niigata University}\\
Niigata, Japan \\
yujun@ie.niigata-u.ac.jp}
\and 
\IEEEauthorblockN{Masaharu Munetomo}
\IEEEauthorblockA{\textit{Information Initiative Center} \\
\textit{Hokkaido University}\\
Sapporo, Japan \\
munetomo@iic.hokudai.ac.jp}
}

\maketitle

\begin{abstract}
Motivated by the potential of large language models (LLMs) as optimizers for solving combinatorial optimization problems, this paper proposes a novel LLM-assisted optimizer (LLMO) to address adversarial robustness neural architecture search (ARNAS), a specific application of combinatorial optimization. We design the prompt using the standard CRISPE framework (i.e., Capacity and Role, Insight, Statement, Personality, and Experiment). In this study, we employ Gemini, a powerful LLM developed by Google. We iteratively refine the prompt, and the responses from Gemini are adapted as solutions to ARNAS instances. Numerical experiments are conducted on NAS-Bench-201-based ARNAS tasks with CIFAR-10 and CIFAR-100 datasets. Six well-known meta-heuristic algorithms (MHAs) including genetic algorithm (GA), particle swarm optimization (PSO), differential evolution (DE), and its variants serve as baselines. The experimental results confirm the competitiveness of the proposed LLMO and highlight the potential of LLMs as effective combinatorial optimizers. The source code of this research can be downloaded from \url{https://github.com/RuiZhong961230/LLMO}.
\end{abstract}


\section{Introduction} \label{sec:1}
In recent years, the remarkable advancements in deep learning have been significantly affected and decelerated by the vulnerability of neural networks (NNs) to adversarial attacks \cite{Liu:23, Hsiung:23}. These attacks, which involve imperceptible perturbations to input data, can lead to highly incorrect predictions \cite{Huang:23}. Consequently, the deployment and configuration of NNs in security-critical applications face serious challenges \cite{Su:19, Wang:23}. As the complexity of these attacks continues to evolve, the imperative to develop robust neural architectures has never been more pressing.

In the meantime, neural architecture search (NAS) has emerged as a potential and powerful tool for designing NNs automatically \cite{Mellor:21, Xue:23, Chu:23, Lu:24}. This technique aims to identify optimal architectures that meet specified performance criteria. However, traditional NAS methods focus primarily on mainstream metrics like accuracy, latency, and model size while the negative effects caused by the adversarial attack are commonly neglected \cite{Devaguptapu:21, Hosseini:21, Bortolussi:24}. Therefore, in the rapid development of the artificial intelligence (AI) community, the adversarial robustness of designed NNs should be considered as a crucial factor, that advances adversarial robustness NAS (ARNAS) techniques. 

Although ARNAS tasks are significantly more complex and challenging than traditional NAS tasks, they are fundamentally combinatorial optimization problems. These can be effectively addressed by approximation optimizers with limited CPU time \cite{Zhong:24_1}. Meanwhile, large language models (LLMs), such as Gemini \cite{Team:23}, GPT \cite{Brown:20, OpenAI:23}, and LLaMA \cite{Hugo:23, Peng:23}, have demonstrated exceptional capabilities in understanding and generating human-like text, solving complex problems, and supporting various domains in AI research \cite{Arun:23, Zhong:24}. The potential of LLMs as optimizers has also been explored: Yang et al. \cite{Yang:24} proposed Optimization by PROmpting (OPRO), where the traveling salesman problem (TSP) is described in natural language and used as input for the LLM to generate new solutions. Liu et al. \cite{Liu:24} introduced LLM-driven EA (LMEA), which uses LLMs as evolutionary combinatorial optimizers with minimal domain knowledge and no additional training required. In LMEA, the LLM selects parent solutions from the current population in each generation, applies search operators to generate offspring, and employs a self-adaptive selection mechanism to ensure the survival of elite solutions. Brahmachary et al. \cite{Shuvayan:24} presented the language-model-based evolutionary optimizer (LEO), a novel population-based optimizer using LLMs to solve numerical optimization problems, including industrial engineering challenges such as supersonic nozzle shape optimization, heat transfer, and windfarm layout optimization. Given these findings, the motivation of this study is to utilize LLMs as combinatorial optimizers to solve ARNAS tasks, potentially accelerating optimization convergence in the search for ARNAS.

This paper proposes a novel LLM-assisted optimizer (LLMO) for addressing ARNAS tasks. Using the standard prompt engineering method CRISPE framework (i.e., Capacity and Role, Insight, Statement, Personality, and Experiment) \cite{Mo:23}, we instruct the LLM Gemini to iteratively search for the optimal architecture under adversarial attacks. Numerical experiments are conducted on the NAS-Bench-201-based search space with CIFAR-10 and CIFAR-100 datasets \cite{Jung:23}. Six well-known discrete meta-heuristic algorithms (MHAs) are employed as competitor algorithms to demonstrate the feasibility and effectiveness of LLMO.

The rest of this paper is organized as follows. Section \ref{sec:2} introduces the related works including the CRISPE framework and ARNAS. Section \ref{sec:3} details our proposed LLMO, Section \ref{sec:4} describes experimental settings and results, we analyze the performance of LLMO in Section \ref{sec:5}, and finally, Section \ref{sec:6} concludes our work.

\section{Related works} \label{sec:2}

\subsection{CRISPE framework}  \label{sec:2.1}
\begin{figure}[htbp]
    \centering
    \includegraphics[width=6cm]{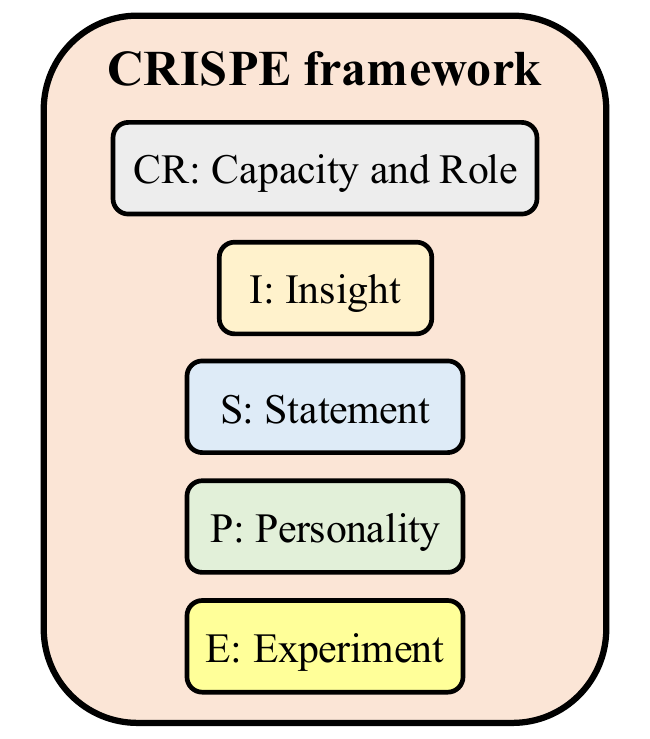}
    \caption{Components in the CRISPE framework.}
    \label{fig:2.1.1}
\end{figure}

Many researchers have recognized the crucial role of prompts in guiding LLMs to produce satisfactory responses \cite{Wei:22, Jules:23}. Consequently, prompt engineering has rapidly emerged as an essential discipline, leading to the development of various techniques such as zero-shot, one-shot, few-shot prompting, and chain-of-thought prompting. This paper adopts the CRISPE framework, a structured prompt engineering approach, to design effective prompts. Here, Fig. \ref{fig:2.1.1} presents the components in the CRISPE framework.
The implications of components are explained as follows:
\begin{itemize}
  \item CR: Define the capacity and role.
  \item I: Provide necessary background or insight.
  \item S: State the core of the prompt.
  \item P: Define the fashion of the LLM's response.
  \item E: Ask for multiple responses.
\end{itemize}

\subsection{Adversarial robustness neural architecture search (ARNAS)}  \label{sec:2.2}
\begin{figure*}[htbp]
    \centering
    \includegraphics[width=16cm]{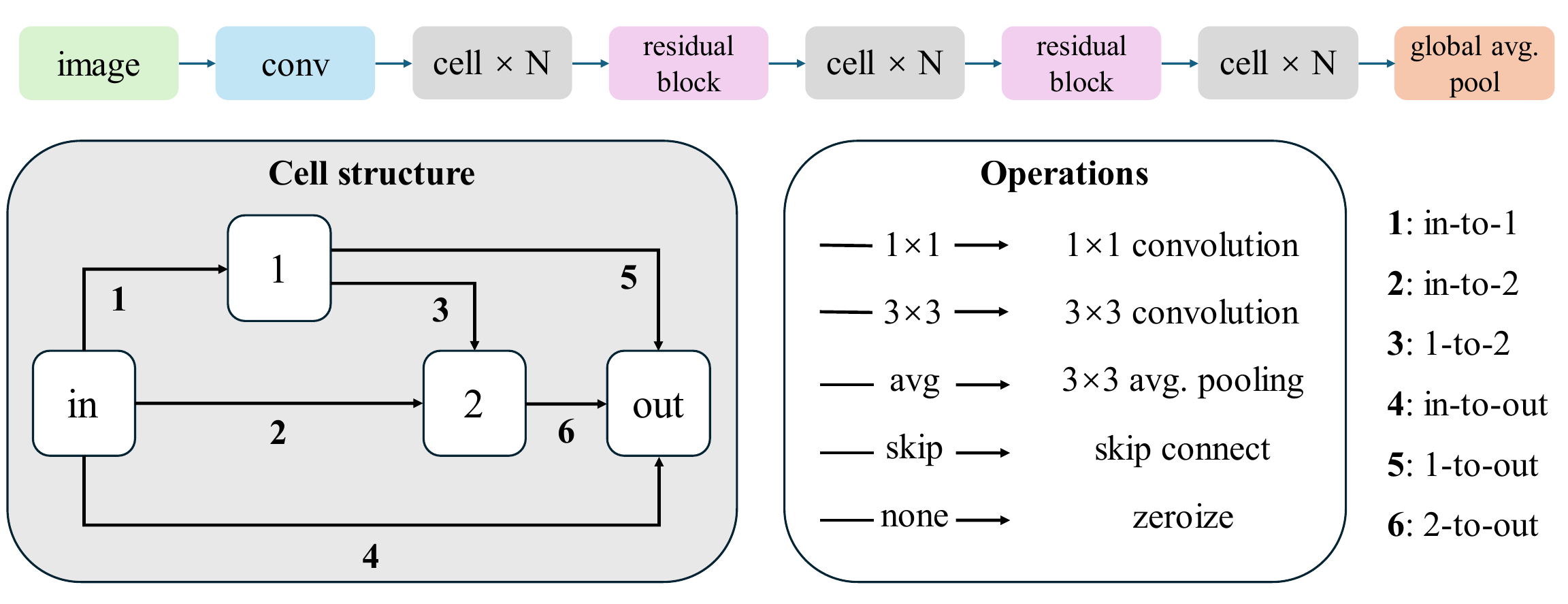}
    \caption{The architecture of NAS-Bench-201-based ARNAS search space.}
    \label{fig:2.2.1}
\end{figure*}

The ARNAS instances presented in \cite{Jung:23} are employed as the benchmark suite in this study, where the structures are demonstrated in Fig. \ref{fig:2.2.1}. A cell in Fig. \ref{fig:2.2.1} consists of four nodes (i.e., feature maps) and six edges (i.e., possible operations). Available operations contain 1x1 convolutions, 3x3 convolutions, 3x3 average pooling, skip connection, and zeroize connection. Consequently, the possible combinations of potential architectures are $5^6=15,625$, among them $6,466$ architectures are non-isomorphic.

Here, four representative adversarial attack methods are adopted: the fast gradient sign method (FGSM) \cite{Ian:15}, projected gradient descent (PGD) \cite{Alexey:17}, adaptive PGD (APGD) \cite{Croce:20}, and square attack \cite{Maksym:20}. Detailed descriptions of these methods can be found in corresponding papers.

\section{Our proposal: LLMO} \label{sec:3}
\begin{figure*}[htbp]
    \centering
    \includegraphics[width=17cm]{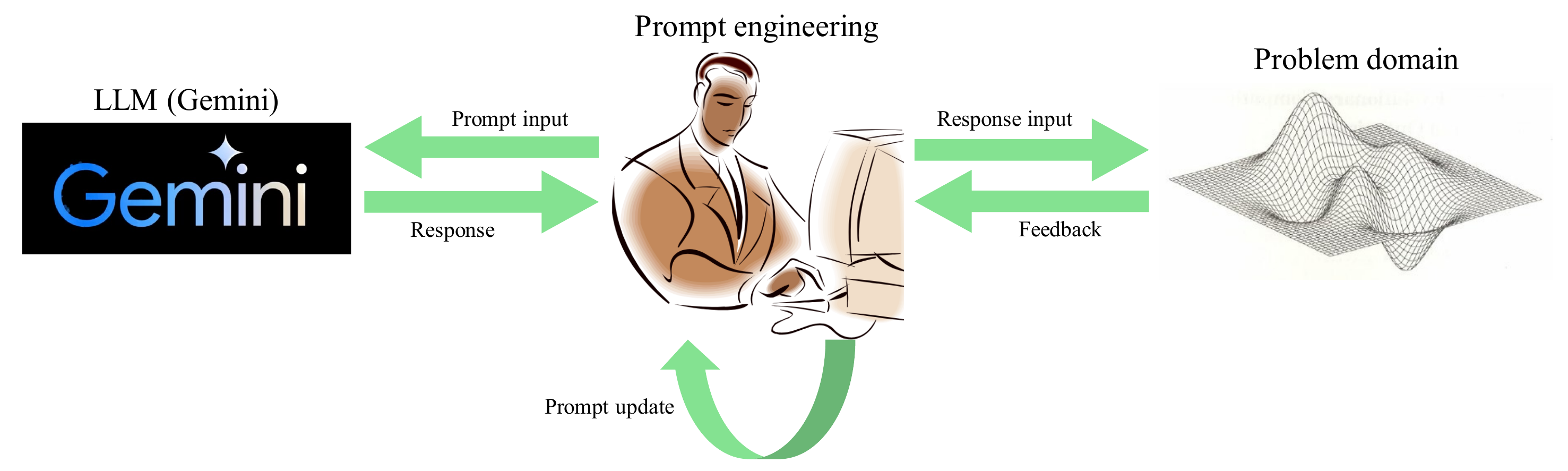}
    \caption{A demonstration of LLMO. }
    \label{fig:3.1}
\end{figure*}

A demonstration of the proposed LLMO is presented in Fig. \ref{fig:3.1}\footnote{The symbol of Gemini is downloaded from \url{https://pixabay.com/illustrations/chip-ai-artificial-intelligence-8530784} as a copyright-free image.}. In the initial step, we design and input the prompt using the CRISPE framework into Gemini. The response from Gemini is then refined as a solution to the ARNAS instance. We utilize the prediction accuracy of the feedback to automatically update the prompt. These processes are repeated until the optimization is complete.

Additionally, the overview of the designed prompt is summarized as follows. The contents within "\{\}" in the prompt will be replaced by the real data during optimization.
\begin{itemize}
  \item CR: Act as a combinatorial optimizer for adversarial robustness neural architecture search.
  \item I: The objective of this task is to maximize the accuracy.
  \item S: There are \{number of operations\} possible operations and \{number of edges\} edges that need to be deployed. You need to specify a \{number of edges\}-bit array where the value in each index is an integer within [0, \{number of operations\}). The current best solution is \{best solution\} with the best accuracy \{best accuracy\}.
  \item P: Not applicable.
  \item E: Give me one solution in the array-like format.
\end{itemize}

In summary, the pseudocode of the proposed LLMO is presented in Algorithm \ref{alg:3.1}. The proposed LLMO offers a significant design advantage: users do not need to understand the working mechanism of Gemini or the specific design of search operators. This means that even amateurs with no prior knowledge of evolutionary computation (EC) and ARNAS can easily use LLMO for optimization. This user-friendly approach ensures easy implementation and accessibility.
\begin{algorithm}
\caption{LLMO}\label{alg:3.1}
\begin{algorithmic}[1]
\Require Max. iteration:$T$
\Ensure Optimum: $\bm{x}^t_{best}$
    \State Randomly initialize the solution $\bm{x}^t_{best}$
    \State Evaluate $\bm{x}^t_{best}$ by ARNAS instance
    \State $t = 0$ 
        \While {$t < T$} 
            \State Construct prompt using CRISPE framework
            \State Input prompt to Gemini
            \State Check the feasibility of the responded solution
            \State Update the current best solution $\bm{x}^t_{best}$
            \State $t \gets t+1$
        \EndWhile
        \State \textbf{return} $\bm{x}^t_{best}$
\end{algorithmic}
\end{algorithm}

\section{Numerical experiments} \label{sec:4}
This section introduces the numerical experiments on ARNAS instances to investigate the performance of LLMO competing with well-known MHAs. Section \ref{sec:4.1} presents the detailed experimental settings and Section \ref{sec:4.2} summarizes the detailed experimental results.

\subsection{Experimental settings} \label{sec:4.1}
We first present the theoretical optimal prediction accuracy of ARNAS instances with the CIFAR-10 and CIFAR-100 datasets in Table \ref{tbl:4.1.1}. Six metaheuristic algorithms (MHAs) are employed as competitor algorithms: genetic algorithm (GA) \cite{John:92}, particle swarm optimization (PSO) \cite{Kennedy:95}, differential evolution (DE) \cite{Storn:96}, evolution strategy with covariance matrix adaptation (CMA-ES) \cite{Hansen:03}, JADE \cite{Zhang:09}, and success-history-adaptive DE (SHADE) \cite{Tanabe:13}. The parameters of these algorithms are summarized in Table \ref{tbl:4.1.2}. The population size of competitor algorithms is fixed at 30 while LLMO is a single solution based optimization approach. The maximum fitness evaluation (FE) of all optimizers is fixed at 30 and 3000, respectively. To alleviate the effect of randomness, each algorithm is implemented in 30 trial runs.
\begin{table}[!ht]
	\scriptsize
	\centering
	\renewcommand\arraystretch{1.3}
	\caption{Summary of the optimal accuracy in the ARNAS benchmark.}
	\label{tbl:4.1.1}
	\resizebox{0.9\columnwidth}{!}{
	\begin{tabular}{cccc}
		\toprule
		Attack method & Optimum in CIFAR-10 & Optimum in CIFAR-100 \\
		\midrule
            Clean (No attack) & 94.6 & 73.6 \\
            FGSM & 69.2 & 29.4 \\
            PGD & 58.8 & 29.8 \\
            APGD & 54.0 & 26.3 \\
            Square & 73.6 & 40.4 \\
		\bottomrule
	\end{tabular}
        }
\end{table}
\begin{table}[htbp]
	\scriptsize
	\centering
	\renewcommand\arraystretch{1.5}
	\caption{The parameters of competitor algorithms.}
	\label{tbl:4.1.2}
	\resizebox{0.9\columnwidth}{!}{
		\begin{tabular}{ccc}
			\toprule
			  Algorithms & Parameters & Value \\
			\midrule
			\multirow{3}{*}{GA} & crossover probability $pc$ & 0.9 \\
			~ & mutation probability $pm$ & 0.01 \\
			~ & selection & tournament \\
                \midrule
                \multirow{3}{*}{PSO} & inertia factor $w$ & 1 \\
                ~ & coefficients $c_1$ and $c_2$ & 2.05 \\
                ~ & max. and min. speed & 2 and -2 \\
			\midrule
			  \multirow{3}{*}{DE} & mutation strategy & DE/cur-to-rand/1/bin \\ 
                ~ & scaling factor $F$ & 0.8 \\
                ~ & crossover rate $Cr$ & 0.9 \\
			\midrule
                CMA-ES & $\sigma$ & 1.3 \\
                \midrule
                JADE & $\mu_{F}$ and $\mu_{Cr}$ & 0.5 and 0.5 \\
			\midrule
			SHADE & $\mu_{F}$ and $\mu_{Cr}$ & 0.5 and 0.5 \\
			\bottomrule
		\end{tabular}
   }
\end{table}

\subsection{Experimental results} \label{sec:4.2}
\begin{figure*}[!ht]
    \centering
    \includegraphics[width=17cm]{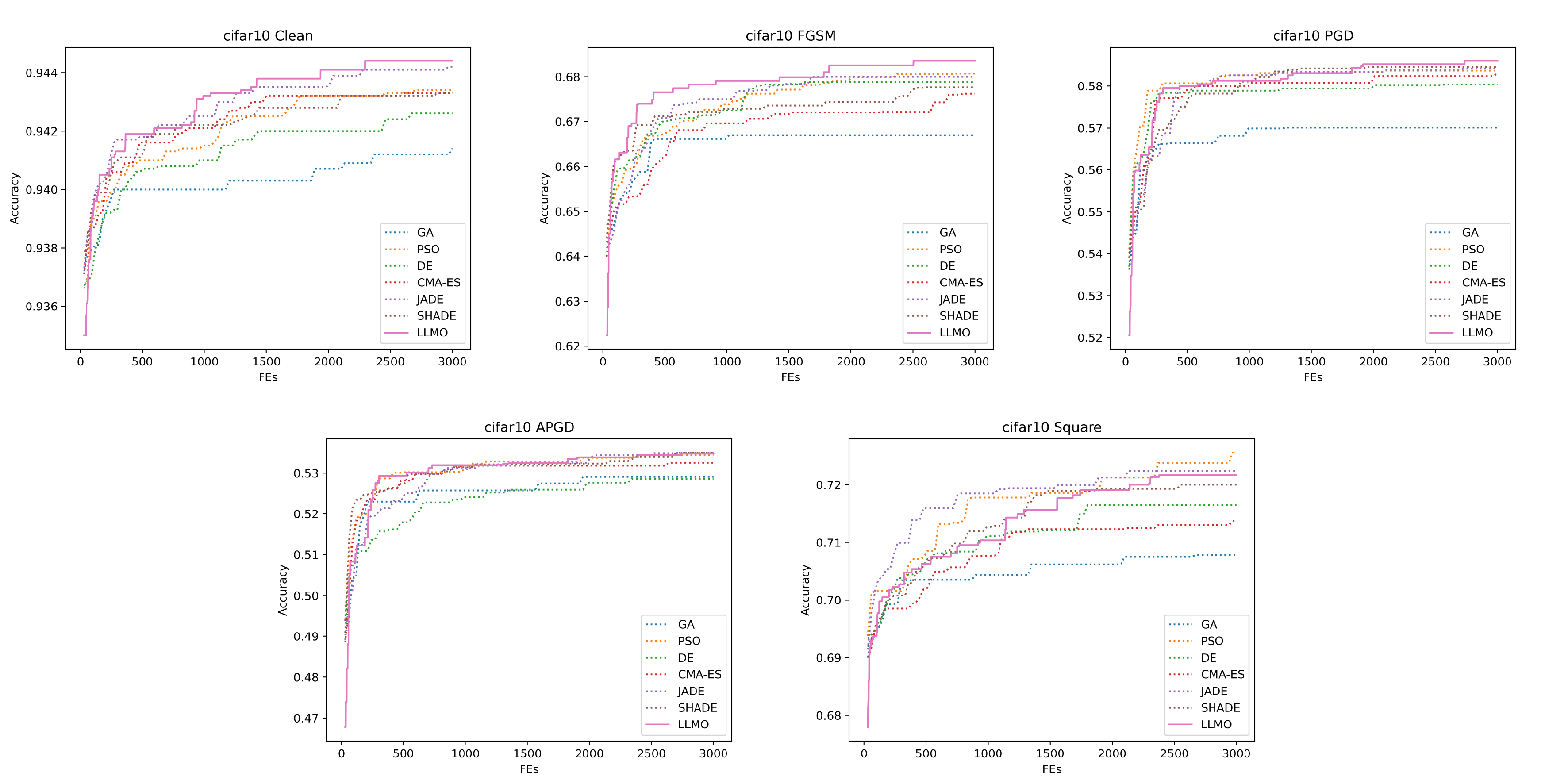}
    \caption{Convergence curves of optimizers for ARNAS on CIFAR-10.}
    \label{fig:4.2.1}
\end{figure*}

\begin{figure*}[!ht]
    \centering
    \includegraphics[width=17cm]{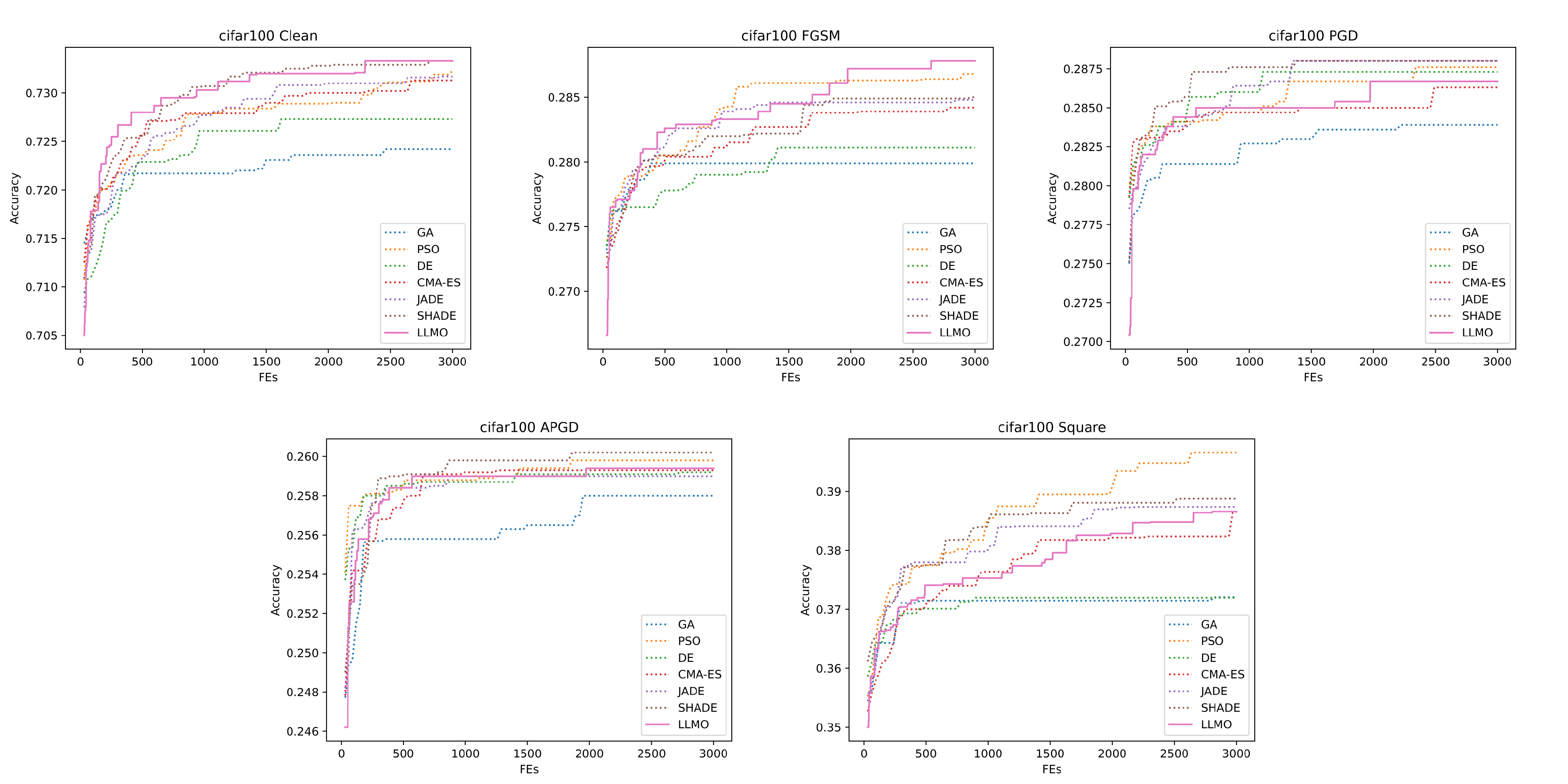}
    \caption{Convergence curves of optimizers for ARNAS on CIFAR-100.}
    \label{fig:4.2.2}
\end{figure*}

The experimental results on the ARNAS tasks are summarized in Table \ref{tbl:4.2.1}, and the convergence curves are presented in Figs. \ref{fig:4.2.1} and \ref{fig:4.2.2}.
\begin{table*}[!ht]
	\scriptsize
	\centering
	\renewcommand\arraystretch{1.5}
	\caption{The experimental results of prediction accuracy in the ARNAS benchmark.}
	\label{tbl:4.2.1}
	\resizebox{1.3\columnwidth}{!}{
		\begin{tabular}{ccccccccccccccccc}
			\toprule
			\multicolumn{2}{c}{Prob.} & GA & PSO & DE & CMA-ES & JADE & SHADE & LLMO \\
			\midrule
			\multirow{5}{*}{CIFAR-10} & Clean & 94.14 & 94.34 & 94.26 & 94.33 & 94.42 & 94.33 & \textbf{94.44} \\
                ~ & FGSM & 66.70 & 68.07 & 67.88 & 67.62 & 68.00 & 67.76 & \textbf{68.35} \\
                ~ & PGD & 57.01 & 58.37 & 58.04 & 58.32 & 58.42 & 58.46 & \textbf{58.60} \\
                ~ & APGD & 52.91 & 53.43 & 52.86 & 53.25 & \textbf{53.50} & 53.49 & 53.47 \\
                ~ & Squares & 70.78 & \textbf{72.56} & 71.65 & 71.37 & 72.24 & 72.00 & 72.17 \\
			\midrule
			\multirow{5}{*}{CIFAR-100} & Clean & 72.42 & 73.23 & 72.73 & 73.13 & 73.17 & 73.33 & \textbf{73.33} \\
                ~ & FGSM & 27.99 & 28.68 & 28.11 & 28.42 & 28.48 & 28.50 & \textbf{28.78} \\
                ~ & PGD & 28.39 & 28.76 & 28.73 & 28.63 & 28.80 & \textbf{28.80} & 28.67 \\
                ~ & APGD & 25.80 & 25.98 & 25.92 & 25.93 & 25.90 & \textbf{26.02} & 25.94 \\
                ~ & Squares & 37.21 & \textbf{39.66} & 37.20 & 38.64 & 38.74 & 38.88 & 38.66 \\
			\bottomrule
		\end{tabular}
	}
\end{table*}

\section{Discussion} \label{sec:5}
The experimental results presented in Table \ref{tbl:4.2.1} and Fig. \ref{fig:4.2.1} confirm the competitiveness of LLMO, particularly in NAS without adversarial attacks and ARNAS with FGSM attacks in both CIFAR-10 and CIFAR-100 datasets. In these four instances, our proposed LLMO outperforms the competitor algorithms, demonstrating the potential and effectiveness of LLM as an optimizer.

Additionally, while GA was originally designed for binary optimization problems, and PSO, DE, CMA-ES, JADE, and SHADE were designed for continuous optimization problems, these algorithms require transfer functions to convert the search domain. The use of transfer functions can lead to different solutions in the original search domain being mapped to identical solutions in the transferred search domain, which deteriorates the quality of constructed offspring individuals and reduces search efficiency. However, this issue does not exist in the proposed LLMO. In the prompt design, the constructed offspring individuals are directly encoded as a {number of edges}-bit array, where the value in each index is an integer within [0, {number of operations}). This direct encoding method for the generation of offspring individuals is more efficient than the approach used by MHAs that rely on transfer functions.

Furthermore, this research reveals the potential of LLMO in solving combinatorial optimization problems, and we believe that it can further adapt to various combinatorial domains such as feature selection \cite{Rui:24}, job scheduling \cite{Meng:24}, and portfolio management problems \cite{Ma:23}.

\section{Conclusion} \label{sec:6}
Motivated by the ability of LLMs to solve combinatorial optimization problems such as TSP, this paper proposes a novel LLM-assisted optimizer (LLMO) to address adversarial robustness neural architecture search (ARNAS) tasks. We design the prompt using the standard CRISPE framework and iteratively refine it during optimization. The experimental results confirm the competitiveness of LLMO and highlight the potential of LLMs as effective optimizers for solving combinatorial optimization problems.

In future research, we will continue to explore the optimization capacity of LLMs in various optimization domains.

\section{Acknowledgement} \label{sec:7}
This work was supported by JSPS KAKENHI Grant Number 21A402 and JST SPRING Grant Number JPMJSP2119.

\bibliographystyle{IEEEtran}
\bibliography{paper}
\end{document}